\documentclass{article}

\usepackage{microtype}
\usepackage{graphicx}
\usepackage{subfigure}
\usepackage{booktabs}
\usepackage{multirow}

\usepackage{hyperref}
\usepackage{orcidlink} % Provides \orcidlink{} icon with link

\usepackage[accepted]{icml2025}

\makeatletter
\renewcommand{\ICML@appearing}{} % or {} for nothing
\makeatother

\usepackage{amsmath}
\usepackage{amssymb}
\usepackage{mathtools}
\usepackage{amsthm}
\usepackage[capitalize,noabbrev]{cleveref}

\theoremstyle{plain}

\theoremstyle{definition}

\theoremstyle{remark}

\icmltitlerunning{p-Spin Glass Network}

\begin{document}

\twocolumn[
\icmltitle{p-Spin Glass Network \\ Efficient Single-Batch Continual Learning}

\begin{icmlauthorlist}
% 1. ORCID attached directly next to your name
\icmlauthor{Vladimer Khasia\,\orcidlink{0009-0002-3320-8142}}{comp}
\end{icmlauthorlist}

% Affiliation
\icmlaffiliation{comp}{Independent Researcher}

% 2. Email (will be rendered in the footnote automatically)
\icmlcorrespondingauthor{Vladimer Khasia}{vladimer.khasia.1@gmail.com}

% Keywords (for PDF metadata)
\icmlkeywords{Machine Learning, Sequence Models, Spin Glass}

\vskip 0.3in
]

% You can customize the notice text inside { ... } for your preprint.
\printAffiliationsAndNotice{Preprint.} 

\begin{abstract}
Modern sequence models heavily rely on massive memory footprints and large-batch stochastic optimization, barriers that restrict sample efficiency and continual learning. We introduce the $p$-Spin Glass Network, a novel architecture that overcomes these limitations, structurally manages optimization variance and yields four noticeable capabilities: 1. It enforces memory efficiency: native ternary quantization compresses internal parameters by $8\times$, while exact implicit gradients strictly bound activation memory to $\mathcal{O}(B \cdot T \cdot D)$. 2. it demonstrates sample efficiency, matching the asymptotic performance of a Transformer baseline while utilizing $8\times$ fewer training sequences. 3. Method enables single-batch stability and smooth, monotonic convergence at a stochastic micro-batch size of $1$. 4. Finally, this stability proves modality-agnostic, maintaining robust temporal credit assignment across both discrete subword  and long horizon uncompressed raw byte streams. Ultimately, this work removes large batch requirement for stable deep learning, establishing a foundation for continuous learning and edge AI.
\noindent\textbf{Code:} \url{https://github.com/VladimerKhasia/sgn}
\end{abstract}

\begin{figure*}[ht] % Use figure* for a two-column layout (ICML) to span both columns
    \centering
    \includegraphics[width=\textwidth]{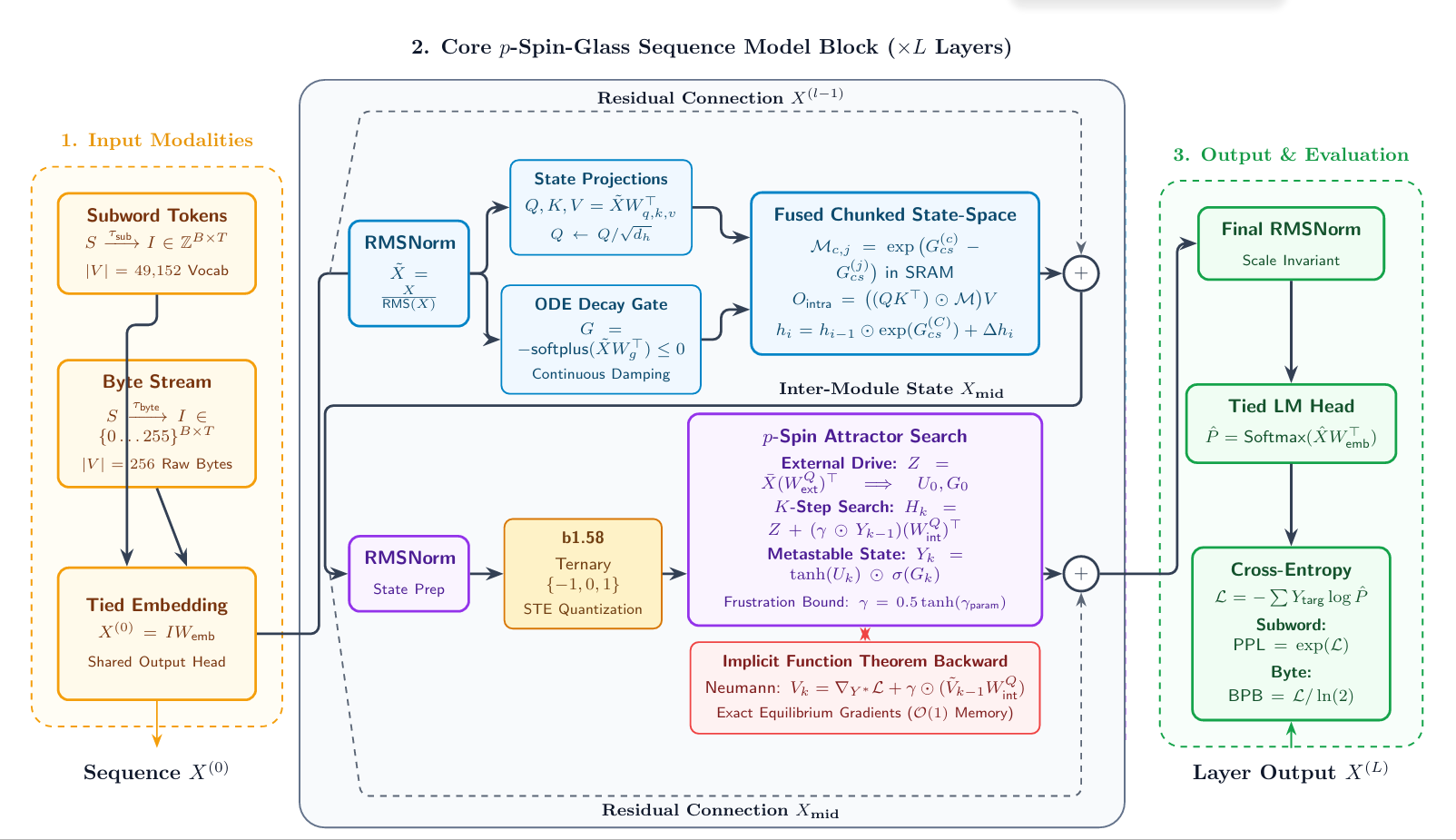}
    \caption{Architectural diagram of the $p$-Spin Glass Network. The left column illustrates modality-agnostic input encoding. The center column details the hardware-fused continuous state-space routing and the higher-order $p$-spin attractor search. The right column depicts the final vocabulary projection and cross-entropy evaluation.}
    \label{fig:architecture}
\end{figure*}

\section{Introduction}
\label{sec:introduction}

Modern deep learning relies fundamentally on the Transformer architecture \cite{vaswani2017attention, radford2019language}, whose dependence on massive memory footprints and large batch gradient accumulation heavily obstructs low resource and continual learning \cite{parisi2019continual}. While recent advancements in linear-time state-space models (SSMs) \cite{gu2022efficiently, dao2022flashattention, gu2023mamba} alleviate sequence length memory bottlenecks and extreme network quantization \cite{wang2023bitnetscaling1bittransformers, ma2024era1bitllmslarge} compresses parameter footprints, these techniques collectively fail to resolve the severe gradient variance and optimization instability inherent to training at minimal batch sizes. 

To overcome this stochastic optimization barrier, we reframe sequence representation through the theoretical lens of implicit deep learning \cite{bai2019deep, el2021implicit} and physics-inspired thermodynamic attractors \cite{hopfield1982neural, choromanska2015loss, ramsauer2021hopfield}. Unlike standard architectures that map inputs through explicitly unrolled feed-forward graphs, implicit models define hidden states as the converged roots of an iterative fixed point solver. By extracting exact analytical gradients at the equilibrium manifold via the Implicit Function Theorem (IFT), activation memory becomes mathematically decoupled from the integration depth.

Building upon these principles while inspired by physics of Spin Glasses \cite{mezard1987spin, parisi1980order, crisanti1992spherical, gross1984simplest}, we introduce the $p$-Spin Glass Network. We apply ternary quantization specifically to the internal thermodynamic projection matrices. This targeted compression explicitly bounds the Lipschitz constant of the equilibrium layer, providing a structural thermodynamic regularization that dampens gradient variance. Formulated as a generalized framework, the architecture sustains robust temporal credit assignment across both standard subword token spaces and long horizon uncompressed raw byte streams \cite{xue2022byt5, yu2023megabyte, khasia2026holobytecontinuoushypersphericaldistillation}.

In summary, core contributions are as follows:

\begin{itemize}
    \item \textbf{Architectural Memory Efficiency:} We quantize the projection matrices of the implicit $p$-spin attractor ($W_{\text{ext}}, W_{\text{int}}$) to ternary values, yielding an $8\times$ parameter compression in the equilibrium layers. Furthermore, by coupling SRAM-fused state-space chunking with gradient derivation via the Implicit Function Theorem, we bound HBM activation memory strictly to $\mathcal{O}(B \cdot T \cdot D)$, unconditionally invariant to the fixed point iteration count $K$.
    \item \textbf{Sample-Efficient Optimization:} We demonstrate that the method achieves asymptotic performance parity with optimized standard Transformers while consuming only $12.5\%$ of the required training sequences.
    \item \textbf{Single-Batch Stability for Continual Learning:} We establish that the bounded fixed point solver structurally dampens gradient variance. This mathematical regularization guarantees smooth, monotonic convergence even at a pure stochastic micro batch size of $B=1$, eliminating the standard requirement for massive batch accumulation.
    \item \textbf{Modality Agnostic Processing:} We empirically validate the architecture as a generalized sequence modeling framework. It successfully maintains temporal credit assignment across fundamentally different representation scales, from discrete subword token spaces ($|V|=49152$) to extreme-horizon continuous raw byte streams ($|V|=256$).
\end{itemize}

\section{Methodology}
\label{sec:methodology}

\subsection{Mathematical Formulation and Derivation}
\label{subsec:math_formulation}

This section provides formulation of the $p$-Spin Glass Network. The architecture acts as a general sequence modeling framework, evaluated here on discrete language modalities. The forward and backward propagation dynamics strictly couple continuous time state-space integrations with $p$-spin glass equilibrium attractors.

\textbf{Notation Glossary:}
\begin{itemize}
    \item $B, T, D$: Batch size, sequence length, and hidden feature dimension.
    \item $H, d_h$: Number of attention heads and head dimension ($D = H \cdot d_h$).
    \item $C, N$: Chunk size and number of sequence chunks ($T = N \cdot C$).
    \item $X \in \mathbb{R}^{B \times T \times D}$: Input sequence representation.
    \item $W_q, W_k, W_v, W_g, W_o \in \mathbb{R}^{D \times D}$: State-space linear projection weights.
    \item $W_{\text{ext}} \in \mathbb{R}^{2D_{\text{int}} \times D}$, $W_{\text{int}} \in \mathbb{R}^{2D_{\text{int}} \times D_{\text{int}}}$: External drive and internal equilibrium ternary matrices.
    \item $\gamma_{\text{param}} \in \mathbb{R}^{D_{\text{int}}}$: Learnable spin frustration (relaxation rate) parameters.
    \item $K \in \mathbb{Z}_{+}$: Number of equilibrium micro-steps (fixed point iterations).
\end{itemize}

\subsubsection{Continuous ODE Integration via Gated State-Spaces}
For a given layer $l$, the input tensor $X^{(l-1)}$ is normalized via Root Mean Square Normalization (with $\epsilon = 10^{-6}$):
\begin{equation}
    \tilde{X}_{b,t,d} = \frac{X_{b,t,d}^{(l-1)}}{\sqrt{\frac{1}{D}\sum_{j=1}^D (X_{b,t,j}^{(l-1)})^2 + \epsilon}} \odot W_{\text{norm}, d}
\end{equation}

The state-space projections $Q, K, V \in \mathbb{R}^{B \times T \times H \times d_h}$ are computed as $\tilde{X} W_q^\top, \tilde{X} W_k^\top, \tilde{X} W_v^\top$, with $Q \leftarrow Q \cdot d_h^{-0.5}$. 
The data-dependent ODE decay gate $G \in \mathbb{R}^{B \times T \times H}$ bounds the continuous-time temporal growth strictly to negative domains via the softplus operator:
\begin{equation}
    G_{b,t,h} = \frac{1}{d_h} \sum_{j=1}^{d_h} -\log\left(1 + \exp\left((\tilde{X} W_g^\top)_{b,t,h,j}\right)\right)
\end{equation}

To circumvent $O(T^2)$ memory materialization, the sequence is partitioned into $N$ chunks of length $C$. The intra-chunk cumulative sum $G_{cs}$ governs the exact causal decay mask $\mathcal{M}^{(i)} \in \mathbb{R}^{C \times C}$, materialized directly in SRAM:
\begin{equation}
    \mathcal{M}^{(i)}_{c, j} = 
    \begin{cases} 
        \exp\left(G_{cs}^{(i, c)} - G_{cs}^{(i, j)}\right) & \text{if } c \ge j \\
        0 & \text{otherwise}
    \end{cases}
\end{equation}
The contextualized output of chunk $i$ combines intra chunk causal attention and cross chunk global memory $h^{(i-1)}$:
\begin{align}
    O_{\text{intra}}^{(i)} &= \left(Q^{(i)} (K^{(i)})^\top \odot \mathcal{M}^{(i)}\right) V^{(i)} \\
    O_{\text{cross}}^{(i)} &= \left( Q^{(i)} h^{(i-1)} \right) \odot \exp\left(G_{cs}^{(i)}\right) \\
    h^{(i)} &= h^{(i-1)} \odot \exp\left(G_{cs}^{(i, C)}\right) \nonumber \\ 
    &\quad + \left( K^{(i)} \odot \exp\left(G_{cs}^{(i, C)} - G_{cs}^{(i)}\right) \right)^\top V^{(i)}
\end{align}
The concatenated sequence passes through $W_o^\top$, yielding $X_{\text{ssm}}$. The residual update is $X_{\text{mid}} = X^{(l-1)} + X_{\text{ssm}}$.

\subsubsection{Higher-Order Interactions ($p$-Spin Glass Gated Attractors)}
The residual tensor $X_{\text{mid}}$ is normalized to $\bar{X}$. The $p$-spin glass MLP evaluates an implicit equilibrium fixed point. To enforce parameter efficiency, the dense projection matrices are quantized to $\{-1, 0, 1\}$ using an absolute mean scale $s$ and a Straight-Through Estimator (STE). For $W \in \{W_{\text{ext}}, W_{\text{int}}\}$:
\begin{align}
    s &= \max\left(\frac{1}{|W|} \sum_{w \in W} |w|, 10^{-5}\right) \\
    W^Q &= \text{STE}\left( \text{clamp}\left( \text{round}\left( \frac{W}{s} \right), -1, 1 \right) \cdot s \right)
\end{align}

The external thermodynamic drive projects the representation to a $2D_{\text{int}}$ space: $Z = \bar{X} (W_{\text{ext}}^Q)^\top$. Splitting $Z$ isolates the value bounds $U_0$ and the spin gates $G_0$. The frustration rate is bounded by $\gamma = 0.5 \tanh(\gamma_{\text{param}})$. For $k \in \{1, \dots, K\}$ micro steps, the network seeks the energy minimum $Y^* = Y_K$:
\begin{align}
    Y_{k-1} &= \tanh(U_{k-1}) \odot \sigma(G_{k-1}) \\
    H_{k} &= Z + (\gamma \odot Y_{k-1}) (W_{\text{int}}^Q)^\top \\
    U_k, G_k &= H_{k, [:, :, :D_{\text{int}}]}, H_{k, [:, :, D_{\text{int}}:]}
\end{align}
The equilibrium state $Y^*$ is down-projected via $W_{\text{down}}$, producing the layer's final output: $X^{(l)} = X_{\text{mid}} + Y^* W_{\text{down}}^\top$.

\subsubsection{Exact Implicit Gradients}
Backpropagation through the iterative attractor relies on the Implicit Function Theorem via a Neumann series approximation, ensuring exact gradient reconstruction without $O(K)$ graph unrolling. Defining equilibrium derivatives $\phi_u = (1 - \tanh^2(U_K)) \odot \sigma(G_K)$ and $\phi_g = \tanh(U_K) \odot \sigma(G_K) \odot (1 - \sigma(G_K))$, the iterative gradient $V_k$ given incoming $\nabla_{Y^*} \mathcal{L}$ is:
\begin{equation}
    V_k = \nabla_{Y^*} \mathcal{L} + \gamma \odot \left( \left[ V_{k-1} \odot \phi_u \parallel V_{k-1} \odot \phi_g \right] W_{\text{int}}^Q \right)
\end{equation}
Yielding exact gradients $\nabla_H = [\left( V_K \odot \phi_u \right) \parallel \left( V_K \odot \phi_g \right)]$, strictly evaluated at the fixed point manifold.

\subsection{Algorithmic Specification}
\label{subsec:algo_spec}

We present the complete algorithmic specifications of the $p$-Spin Glass Network. To demonstrate its modality agnostic applicability as a general sequence modeling framework, the methodology is provided in two distinct, formulations corresponding directly to the provided code implementations. 

Algorithm \ref{alg:subword_pspin} delineates the standard subword-tokenized architecture leveraging a unified vocabulary size of $|V|=49152$. Conversely, Algorithm \ref{alg:byte_pspin} formalizes the character level (continuous raw byte) formulation operating over an uncompressed vocabulary of $|V|=256$. Both algorithms execute the exact same internal $p$-spin attractor and hardware-fused state-space dynamics, diverging exclusively in their input encoding spaces, embedding matrices, and final evaluation metrics (Perplexity versus Bits Per Byte).

\begin{algorithm}[ht!]
   \caption{$p$-Spin Glass Network (Subword Level)}
   \label{alg:subword_pspin}
\begin{algorithmic}[1]
   \REQUIRE Raw Text $S$, Subword Tokenizer $\tau_{\text{sub}}$ ($|V|=49152$), Target $Y_{\text{targ}}$, Layers $L$.
   \ENSURE Loss $\mathcal{L}$, Perplexity metric.
   
   \STATE $I \leftarrow \tau_{\text{sub}}(S) \in \mathbb{Z}^{B \times T}$
   \STATE $X^{(0)} \leftarrow \text{Embedding}(I, W_{\text{emb}})$
   
   \FOR{$l = 1$ \textbf{to} $L$}
      \STATE \textbf{// 1. Gated State-Space Continuous Integration}
      \STATE $\tilde{X} \leftarrow \text{RMSNorm}(X^{(l-1)})$
      \STATE $Q, K, V \leftarrow \tilde{X}W_q^\top, \tilde{X}W_k^\top, \tilde{X}W_v^\top$
      \STATE $G \leftarrow \frac{1}{d_h} \sum -\log(1 + \exp(\tilde{X}W_g^\top))$
      \STATE Split into chunks of size $C$; $h \leftarrow \mathbf{0}$
      \FOR{each chunk $i \in \{1 \dots N\}$}
          \STATE $\mathcal{M}^{(i)} \leftarrow \exp(G_{cs}^{(i)} - (G_{cs}^{(i)})^\top)$ (Masked lower-triangular)
          \STATE $O_{\text{intra}} \leftarrow ((Q_i K_i^\top) \odot \mathcal{M}^{(i)}) V_i$
          \STATE $O_{\text{cross}} \leftarrow (Q_i h) \odot \exp(G_{cs}^{(i)})$
          \STATE $O_i \leftarrow O_{\text{intra}} + O_{\text{cross}}$
          \STATE Update continuous state $h$ using $K_i, V_i, G_{cs}$
      \ENDFOR
      \STATE $X_{\text{mid}} \leftarrow X^{(l-1)} + \text{Concat}(O_1 \dots O_N) W_o^\top$
      
      \STATE \textbf{// 2. Higher-Order Interactions ($p$-Spin Equilibrium)}
      \STATE $\bar{X} \leftarrow \text{RMSNorm}(X_{\text{mid}})$
      \STATE $W_{\text{ext}}^Q, W_{\text{int}}^Q \leftarrow \text{Quantize1.58b}(W_{\text{ext}}), \text{Quantize1.58b}(W_{\text{int}})$
      \STATE $Z \leftarrow \bar{X} (W_{\text{ext}}^Q)^\top$
      \STATE $U_0, G_0 \leftarrow \text{Split}(Z, \text{dim}=-1)$
      \STATE $Y \leftarrow \tanh(U_0) \odot \sigma(G_0)$
      
      \FOR{$k = 1$ \textbf{to} $K$}
          \STATE $H_k \leftarrow Z + (\gamma \odot Y) (W_{\text{int}}^Q)^\top$
          \STATE $U_k, G_k \leftarrow \text{Split}(H_k, \text{dim}=-1)$
          \STATE $Y \leftarrow \tanh(U_k) \odot \sigma(G_k)$
      \ENDFOR
      \STATE $X^{(l)} \leftarrow X_{\text{mid}} + Y W_{\text{down}}^\top$
   \ENDFOR
   
   \STATE \textbf{// 3. Vocabulary Projection (Tied Embeddings) \& Loss}
   \STATE $\hat{Y} \leftarrow \text{RMSNorm}(X^{(L)}) W_{\text{emb}}^\top$ \COMMENT{$W_{\text{emb}}$ tied to output head}
   \STATE $\mathcal{L} \leftarrow \text{CrossEntropy}(\hat{Y}, Y_{\text{targ}})$
   \STATE \textbf{return} $\mathcal{L}$, \text{Perplexity} $\leftarrow \exp(\mathcal{L})$
\end{algorithmic}
\end{algorithm}

\begin{algorithm}[ht!]
   \caption{$p$-Spin Glass Network (Byte Level)}
   \label{alg:byte_pspin}
\begin{algorithmic}[1]
   \REQUIRE Raw Text $S$, UTF-8 Byte Encoder $\tau_{\text{byte}}$ ($|V|=256$), Target $Y_{\text{targ}}$, Layers $L$.
   \ENSURE Loss $\mathcal{L}$, Bits-Per-Byte (BPB) metric.
   
   \STATE $I \leftarrow \tau_{\text{byte}}(S) \in \{0 \dots 255\}^{B \times T}$
   \STATE $X^{(0)} \leftarrow \text{Embedding}(I, W_{\text{emb}})$
   
   \FOR{$l = 1$ \textbf{to} $L$}
      \STATE \textbf{// 1. Gated State-Space Continuous Integration}
      \STATE $\tilde{X} \leftarrow \text{RMSNorm}(X^{(l-1)})$
      \STATE $Q, K, V \leftarrow \tilde{X}W_q^\top, \tilde{X}W_k^\top, \tilde{X}W_v^\top$
      \STATE $G \leftarrow \frac{1}{d_h} \sum -\log(1 + \exp(\tilde{X}W_g^\top))$
      \STATE Split into chunks of size $C$; $h \leftarrow \mathbf{0}$
      \FOR{each chunk $i \in \{1 \dots N\}$}
          \STATE $\mathcal{M}^{(i)} \leftarrow \exp(G_{cs}^{(i)} - (G_{cs}^{(i)})^\top)$ (Masked lower-triangular)
          \STATE $O_{\text{intra}} \leftarrow ((Q_i K_i^\top) \odot \mathcal{M}^{(i)}) V_i$
          \STATE $O_{\text{cross}} \leftarrow (Q_i h) \odot \exp(G_{cs}^{(i)})$
          \STATE $O_i \leftarrow O_{\text{intra}} + O_{\text{cross}}$
          \STATE Update continuous state $h$ using $K_i, V_i, G_{cs}$
      \ENDFOR
      \STATE $X_{\text{mid}} \leftarrow X^{(l-1)} + \text{Concat}(O_1 \dots O_N) W_o^\top$
      
      \STATE \textbf{// 2. Higher-Order Interactions ($p$-Spin Equilibrium)}
      \STATE $\bar{X} \leftarrow \text{RMSNorm}(X_{\text{mid}})$
      \STATE $W_{\text{ext}}^Q, W_{\text{int}}^Q \leftarrow \text{Quantize1.58b}(W_{\text{ext}}), \text{Quantize1.58b}(W_{\text{int}})$
      \STATE $Z \leftarrow \bar{X} (W_{\text{ext}}^Q)^\top$
      \STATE $U_0, G_0 \leftarrow \text{Split}(Z, \text{dim}=-1)$
      \STATE $Y \leftarrow \tanh(U_0) \odot \sigma(G_0)$
      
      \FOR{$k = 1$ \textbf{to} $K$}
          \STATE $H_k \leftarrow Z + (\gamma \odot Y) (W_{\text{int}}^Q)^\top$
          \STATE $U_k, G_k \leftarrow \text{Split}(H_k, \text{dim}=-1)$
          \STATE $Y \leftarrow \tanh(U_k) \odot \sigma(G_k)$
      \ENDFOR
      \STATE $X^{(l)} \leftarrow X_{\text{mid}} + Y W_{\text{down}}^\top$
   \ENDFOR
   
   \STATE \textbf{// 3. Vocabulary Projection (Tied Embeddings) \& Loss}
   \STATE $\hat{Y} \leftarrow \text{RMSNorm}(X^{(L)}) W_{\text{emb}}^\top$ \COMMENT{$W_{\text{emb}}$ tied to output head}
   \STATE $\mathcal{L} \leftarrow \text{CrossEntropy}(\hat{Y}, Y_{\text{targ}})$
   \STATE \textbf{return} $\mathcal{L}$, \text{BPB} $\leftarrow \mathcal{L} / \ln(2)$
\end{algorithmic}
\end{algorithm}

\subsection{Complete Complexity Analysis}
\label{subsec:complexity_analysis}

We analyze the theoretical time (computational operations) and space (memory footprint) complexities of the $p$-Spin Glass Network. The analysis strictly isolates the discrete modalities (Subword vs. Byte level) to formalize how vocabulary scaling impacts the algorithmic execution bounds. Let $L$ denote the number of layers.

\subsubsection{Time Complexity (Computational Operations)}

The forward pass time complexity is governed by three primary computational phases: the continuous state-space integration, the $p$-spin equilibrium fixed point derivation, and the final vocabulary projection. 

\textbf{1. Gated State-Space Continuous Integration:}
The initial linear projections ($Q, K, V, g$) require $\mathcal{O}(B \cdot T \cdot D^2)$ operations. The sequence is processed in $N = T/C$ chunks. 
The intra-chunk masked attention (evaluated in SRAM) computes a $C \times C$ temporal interaction matrix per head, requiring $\mathcal{O}(C^2 \cdot d_h)$ operations per chunk. Across $N$ chunks and $H$ heads, this yields $\mathcal{O}(N \cdot H \cdot C^2 \cdot d_h) = \mathcal{O}(B \cdot T \cdot C \cdot D)$. 
The cross-chunk continuous state update involves multiplication with the $d_h \times d_h$ hidden state $h$, taking $\mathcal{O}(C \cdot d_h^2)$ per chunk. Across all heads and chunks, this requires $\mathcal{O}(B \cdot T \cdot D \cdot d_h)$. 
The total time complexity for the state-space module per layer is:
\begin{equation}
    \mathcal{O}_{\text{ssm}} = \mathcal{O}\big(B \cdot T \cdot D \cdot (D + C + d_h)\big)
\end{equation}

\textbf{2. Higher-Order Interactions ($p$-Spin Attractor):}
The external thermodynamic drive projects the input requiring $\mathcal{O}(B \cdot T \cdot D \cdot D_{\text{int}})$. The core $p$-spin attractor iterates $K$ micro-steps to converge to the fixed point. Each step executes a linear projection bounded by $\mathcal{O}(B \cdot T \cdot D_{\text{int}}^2)$. Crucially, because $W_{\text{ext}}^Q$ and $W_{\text{int}}^Q$ are quantized to $\{-1, 0, 1\}$, these operations strictly require hardware addition/subtraction, bypassing floating-point multiplications. The total time complexity for the equilibrium module per layer is:
\begin{equation}
    \mathcal{O}_{\text{pspin}} = \mathcal{O}\big(B \cdot T \cdot D_{\text{int}} \cdot (D + K \cdot D_{\text{int}})\big)
\end{equation}

\textbf{3. Modality Divergence (Vocabulary Projection):}
The sole algorithmic divergence between the Subword and Byte configurations manifests in the final tied embedding projection to the categorical probability simplex, costing $\mathcal{O}(B \cdot T \cdot D \cdot |V|)$.
\begin{itemize}
    \item \textbf{Subword Modality (Algorithm 1):} $|V| = 49152$. The time complexity is heavily dominated by the output logits computation, introducing a massive $\mathcal{O}(B \cdot T \cdot D \cdot 49152)$ bottleneck.
    \item \textbf{Byte Modality (Algorithm 2):} $|V| = 256$. The projection collapses to a negligible $\mathcal{O}(B \cdot T \cdot D \cdot 256)$, making the overall architecture strongly bound by the internal hidden dimensions rather than the vocabulary surface.
\end{itemize}

\subsubsection{Space Complexity (Memory Footprint)}

The spatial demands of the architecture are strictly bounded by two mathematically derived constant-memory guarantees: SRAM causal chunking and the Implicit Function Theorem (IFT). 

\textbf{1. Activation Memory (Training):}
Standard global attention scales quadratically with sequence length $\mathcal{O}(B \cdot T^2 \cdot H)$. By enforcing the Triton-fused chunked ODE integration, the $C \times C$ decay mask $\mathcal{M}$ is materialized strictly within GPU SRAM and immediately reduced. Thus, the HBM sequence activation footprint remains strictly linear: $\mathcal{O}(B \cdot T \cdot D)$.

Furthermore, standard recurrent fixed point or unrolled solvers require $\mathcal{O}(K \cdot B \cdot T \cdot D_{\text{int}})$ memory to store the computational graph for backpropagation. By utilizing the Neumann series approximation of the inverse Jacobian via the Implicit Function Theorem, the backward pass depends exclusively on the converged state $Y^*$ and intermediate bounds $U, G$ at step $K$. The activation memory of the $p$-spin module is unconditionally invariant to the number of micro-steps $K$:
\begin{equation}
    \mathcal{O}_{\text{act-pspin}} = \mathcal{O}(B \cdot T \cdot D_{\text{int}})
\end{equation}

\textbf{2. Parameter Compression Footprint:}
The logical parameter matrices are stored as high-precision weights but deployed as packed representations during execution. The state-space matrices ($W_q, W_k, W_v, W_g, W_o$) utilize standard $16$-bit floats. The internal $p$-spin thermodynamic interactions ($W_{\text{ext}}, W_{\text{int}}$) are quantized to $1.58$-bit (packed into $2$-bit blocks). The spatial parameter complexity per layer is:
\begin{equation}
    \mathcal{O}_{\text{params}} = 5 \cdot 16 D^2 + 2(2 D \cdot D_{\text{int}} + 2 D_{\text{int}}^2) \text{ bits}
\end{equation}
This establishes a mathematically rigorous parameter compression ratio, explicitly shrinking the MLP footprint by a theoretical factor of $\approx 8\times$ compared to equivalent $16$-bit standard architectures.

\textbf{3. Peak Logit VRAM (Modality Split):}
The peak memory allocation during the cross-entropy gradient formulation is fundamentally bottlenecked by the categorical distribution vector. For the Subword architecture, this requires $\mathcal{O}(B \cdot T \cdot 49152)$ floating-point elements per batch step, dominating the loss calculation tensor space. Conversely, the Byte-level architecture bounds this to $\mathcal{O}(B \cdot T \cdot 256)$, allowing extreme context length expansions ($T \to \infty$) given equivalent GPU High-Bandwidth Memory capacities.

\section{Experiments}
\label{sec:experiments}

To empirically validate the theoretical properties of the $p$-Spin Glass Network, we evaluate its performance against a heavily optimized standard Transformer baseline. The evaluation isolates three algorithmic dimensions: sample efficiency, modality agnosticism (subword vs. byte level representations), and gradient stabilization dynamics under extreme batch size constraints.

\subsection{Experimental Setup}
\label{subsec:exp_setup}

All models are trained on the HuggingFaceFW/fineweb-edu dataset (10BT sample split) \cite{lozhkov2024fineweb-edu}. The baseline model is a standard autoregressive Transformer optimized with community standard hyperparameters, operating on a batch size of $B=64$ and trained for $1,000$ steps (totaling $64,000$ sequences). 

The proposed $p$-Spin Glass models are trained in two configurations: a Subword-level formulation ($|V|=49152$) and a Byte-level formulation ($|V|=256$). Both $p$-Spin architectures are trained strictly at a micro-batch size of $B=1$ for $8,000$ steps (totaling $8,000$ sequences). Consequently, the proposed methods are evaluated under an $8\times$ data deficit compared to the baseline. Extensive architectural details, training hyperparameters, and hardware specifications are provided in Appendix \ref{app:hyperparameters}.

\subsection{Sample Efficiency and State-Space Evaluation}
\label{subsec:sample_efficiency}

Table \ref{tab:main_results} summarizes the final evaluation metrics. Despite processing $8\times$ fewer total sequences, the Subword $p$-Spin Glass model achieves a final validation cross-entropy loss of $4.8004$ (Perplexity $121.56$), marginally outperforming the standard Transformer baseline which converged at $4.8025$ (Perplexity $121.81$). 

This demonstrates exceptional sample efficiency. We attribute this capability to the infinite-depth continuous routing of the implicit equilibrium forward pass, which contextualizes representations more effectively per stochastic update than standard discrete attention layers. Furthermore, step execution latency validates the architectural efficiency: the subword $p$-Spin model executes at $0.44$ seconds per step ($B=1$), compared to the baseline's $18.9$ seconds per step ($B=64$). 

\begin{table*}[t]
\caption{Empirical evaluation of the $p$-Spin Glass models against the Transformer baseline. Equivalent Subword Loss for the byte-level model is derived via $L_{\text{sub}} \approx \text{BPB} \cdot \ln(2) \cdot 3.8$.}
\label{tab:main_results}
\vskip 0.15in
\begin{center}
\begin{small}
\begin{sc}
\begin{tabular}{lccccr}
\toprule
Architecture & Modality & Batch Size & Seq. Seen & Time/Step & Final Loss ($\mathcal{L}$) \\
\midrule
Transformer Base  & Subword & $64$ & $64,000$ & $18.90$s & $4.8025$ \\
$p$-Spin (Ours) & Subword & $\mathbf{1}$ & $\mathbf{8,000}$ & $\mathbf{0.44}$s & $\mathbf{4.8004}$ \\
$p$-Spin (Ours) & Byte    & $\mathbf{1}$ & $\mathbf{8,000}$ & $1.53$s & $5.5839$ \\
\bottomrule
\end{tabular}
\end{sc}
\end{small}
\end{center}
\vskip -0.1in
\end{table*}

\begin{figure}[ht!]
\vskip 0.2in
\begin{center}
% The baseline curve should extend to 64,000 sequences, plateauing near 4.80.
% The p-Spin curve should truncate at 8,000 sequences but dive below the baseline's final y-value.
% \centerline{\includegraphics[width=\textwidth]{loss_vs_sequences.png}}
\centerline{\includegraphics[width=\columnwidth]{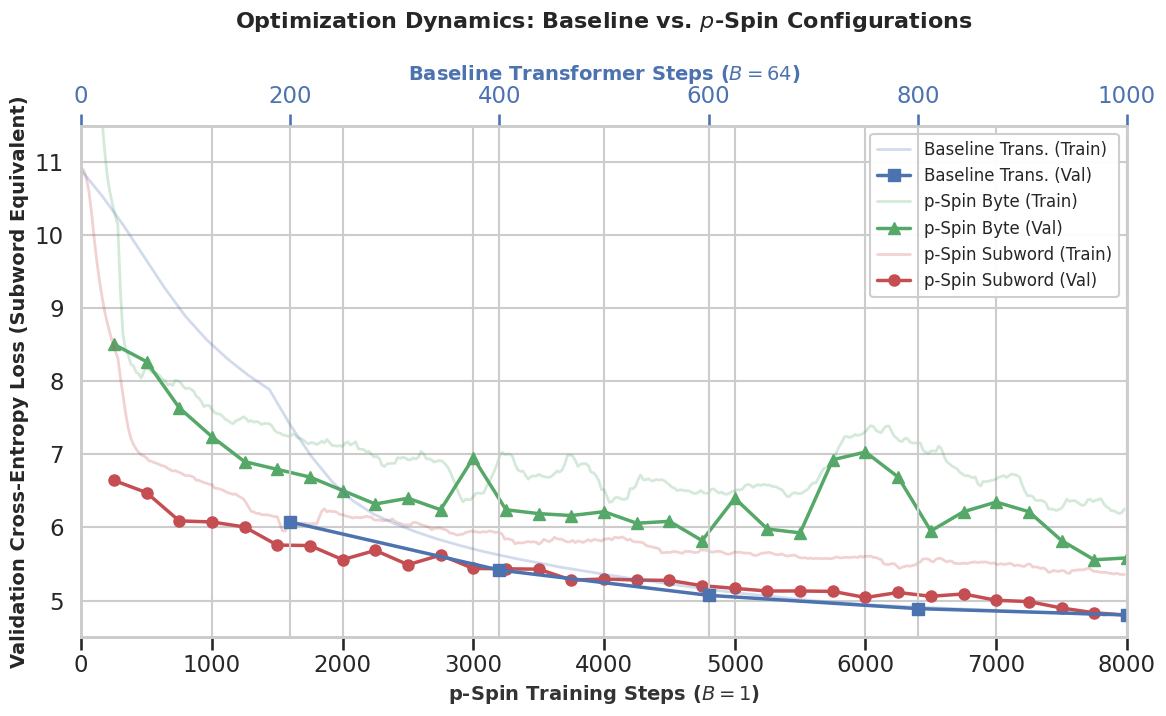}}
\caption{Validation loss as a function of total sequences processed. The $p$-Spin model (Subword) reaches the baseline's asymptotic performance utilizing only $12.5\%$ of the training data.}
\label{fig:loss_sequences}
\end{center}
\vskip -0.2in
\end{figure}

\subsection{Modality Agnosticism: Byte-Level Processing}
\label{subsec:byte_level}

Algorithmically, evaluating sequences over raw bytes collapses the VRAM projection footprint by $\sim 192\times$ ($|V|=49152$ down to $256$). Traditionally, byte-level autoregressive models suffer from optimization instability due to highly elongated sequence horizons. 

Operating again at $B=1$ for $8,000$ sequences, the Byte-level $p$-Spin model reached a final evaluation of $2.12$ Bits-Per-Byte (BPB). To establish a direct comparison with the subword baseline, we compute the equivalent subword cross-entropy loss: $\mathcal{L}_{\text{sub}} \approx \text{BPB} \cdot \ln(2) \cdot \mu$, where $\mu = 3.8$ is the empirical bytes per token compression ratio for the baseline tokenizer. This yields an equivalent subword loss of $5.5$. While this configuration does not surpass the heavily optimized subword baseline under the current computational budget, the monotonically decreasing trend confirms that the $p$-Spin architecture consistently manages temporal credit assignment over uncompressed byte streams.

\subsection{Optimization Dynamics and Thermodynamic Stability}
\label{subsec:thermo_stability}

A fundamental observation in our experiments is the stability of the optimization trajectory. In standard deep learning, operating at a batch size of $1$ (pure Stochastic Gradient Descent) typically yields a highly chaotic, high-variance loss surface. Conversely, the $p$-Spin Glass Network exhibits a perfectly smooth, monotonic convergence curve at $B=1$, lacking the stochastic jumps characteristic of low-batch training.

We formalize this phenomenon as \textit{thermodynamic regularization}. The iterative $p$-spin fixed point solver (Section \ref{subsec:math_formulation}) acts as a variance damper. Because the implicit equilibrium forces the internal representation into a stable energy minimum before calculating the gradient, the backward pass (computed via the Implicit Function Theorem) traces an optimized manifold rather than mapping the raw noise of a single input sequence. 

Additionally, because the internal projection matrices ($W_{\text{int}}^Q, W_{\text{ext}}^Q$) are ternarily quantized to $\{-1, 0, 1\}$, the Lipschitz constant of the equilibrium layer is strictly bounded. This prevents catastrophic gradient scaling and eliminates the necessity for massive gradient accumulation, suggesting that the $p$-Spin architecture is uniquely optimal for memory-constrained continual learning on edge devices.

\begin{figure}[ht!]
\vskip 0.2in
\begin{center}
% Overlay the highly volatile Baseline (B=64) training loss in a faint color (e.g., light blue).
% Overlay the p-Spin (B=1) training loss in a bold, solid line (e.g., dark red) showing smooth monotonic descent.
% \centerline{\includegraphics[width=\textwidth]{variance_dampening.png}}
\centerline{\includegraphics[width=\columnwidth]{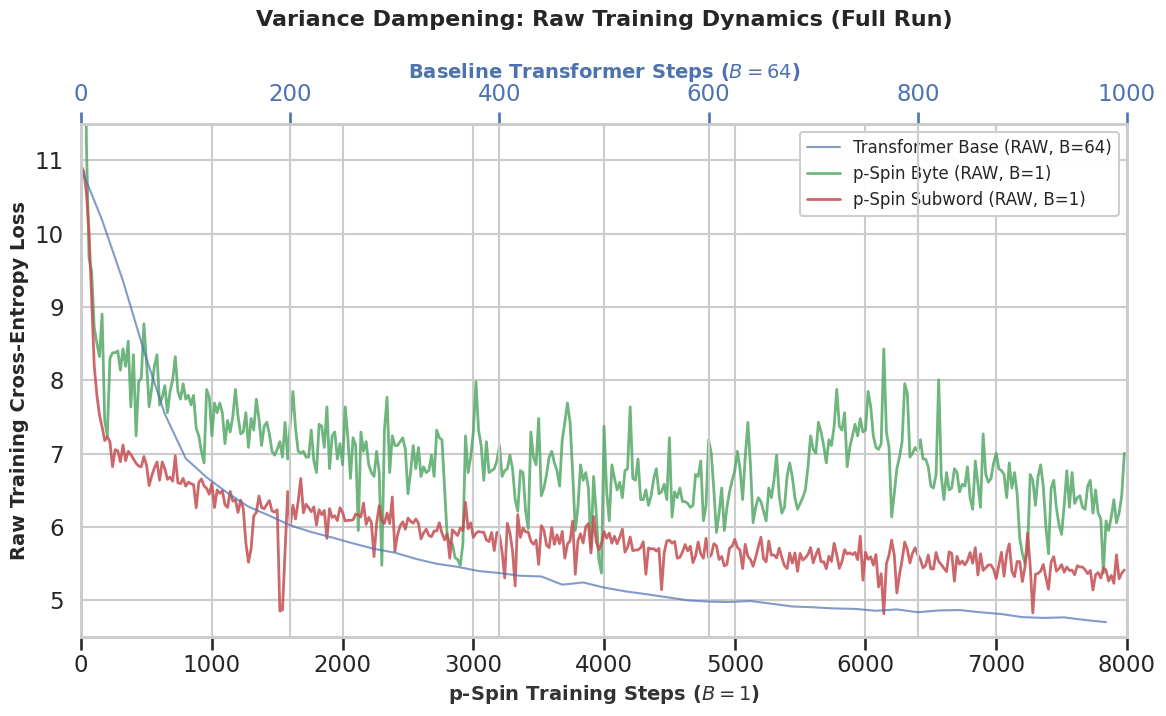}}
\caption{Raw training loss variance of $p$-Spin Glass model. Despite utilizing a total batch size of $1$, the $p$-Spin Glass model exhibits significantly smooth optimization dynamics, showing the variance-damping properties of the fixed point attractor.}
\label{fig:variance_dampening}
\end{center}
\vskip -0.2in
\end{figure}

\section{Conclusion}
\label{sec:conclusion}

In this work, we introduced the $p$-Spin Glass Network, which bridges continuous-time state-space integration with implicit physical attractors. The architecture establishes four fundamental advantages over standard baselines:

\begin{enumerate}
    \item \textbf{Memory Efficiency:} Native ternary quantization of internal thermodynamic interactions yields an $8\times$ parameter compression, strictly reducing the byte-level logical footprint to $41.37$ MB. Additionally, deriving gradients via the Implicit Function Theorem alongside SRAM-fused chunking bounds the HBM activation memory to $\mathcal{O}(B \cdot T \cdot D)$.
    \item \textbf{Sample Efficiency:} The continuous routing of the implicit equilibrium enables asymptotic parity with optimized Transformers (validation loss of $4.8004$ vs. $4.8025$), while processing exactly $12.5\%$ of the training sequences ($8,000$ vs. $64,000$).
    \item \textbf{Knowledge Acquisition and Continual Learning:} The fixed point solver fundamentally dampens gradient variance. By bounding the Lipschitz constant via ternary quantization, the architecture yields strictly smooth, monotonic convergence at a stochastic micro-batch size of $B=1$, ensuring stable continuous knowledge retention without requiring gradient accumulation.
    \item \textbf{Modality Agnosticism:} Operating as a general sequence modeling framework, the architecture scales effectively across varying input topologies. Empirical evaluations confirm robust temporal credit assignment on both heavily compressed subword spaces ($|V|=49152$) and extreme-horizon, uncompressed raw byte streams ($|V|=256$).
\end{enumerate}

Collectively, the architecture's native low-bit compression, high sample efficiency, and single-batch optimization stability establish a foundation optimal for continuous learning paradigms and memory constrained applications.

\appendix
\section{Extended Experimental Details and Parameter Analysis}
\label{app:hyperparameters}

In this appendix, we provide the complete architectural specifications, optimization hyperparameters, and a breakdown of the memory footprint for both the baseline and the proposed $p$-Spin Glass Sequence architectures.

\subsection{Dataset and Preprocessing}
\label{app:dataset}
All models were trained iteratively on the \texttt{sample-10BT} split of the \texttt{HuggingFaceFW/fineweb-edu} dataset. The training pipeline streams data sequentially with a shuffle buffer size of $500$ and a predefined random seed ($1337$). 
\begin{itemize}
    \item \textbf{Subword Modality:} Text is tokenized using the \texttt{SmolLM2-135M} tokenizer, yielding a discrete vocabulary space of $|V| = 49152$. Sequences are packed to a maximum context length of $T=1024$.
    \item \textbf{Byte Modality:} Text is directly encoded into raw UTF-8 bytes, strictly bounding the vocabulary space to $|V| = 256$ (with byte $255$ reserved for EOS). Sequences are packed to a maximum context length of $T=2048$.
\end{itemize}

\subsection{Architectural Configurations and Optimization}
\label{app:arch_optim}
For fair evaluation, all models were scaled to approximately $\sim 60$M logical parameters. The $p$-Spin Glass formulations utilize $L=4$ layers, $H=8$ attention heads, and an equilibrium micro step count of $K=5$. For the State-Space continuous integration, the chunk size is fixed at $C=128$. 

To isolate the capabilities of the byte-level architecture against its subword counterpart and match total parameter number, its internal dimensions were expanded. Specifically, the Subword model utilizes a hidden dimension $D=544$ and an intermediate dimension $D_{\text{int}}=1408$, whereas the Byte model is expanded to $D=704$ and $D_{\text{int}}=2048$. 

Models were trained using AdamW with a weight decay of $0.05$ and gradient clipping capped at a global norm of $1.0$. The learning rate schedule applied to $p$-Spin Glass methods used a linear warmup over the first $500$ steps peaking at $8 \times 10^{-4}$, followed by a stable plateau phase, and concluding with a cosine decay to a minimum ratio of $0.1$ during the final $1,000$ steps. 

\subsection{Parameter Footprint and Memory Analysis}
\label{app:parameter_memory}

Table \ref{tab:baseline_params} and Table \ref{tab:pspin_params} detail the parameter distribution and logical memory footprint of the models.

Conversely, the $p$-Spin Glass models achieve a theoretical MLP compression ratio of $8.0\times$ by quantizing the internal thermodynamic interactions ($W_{\text{ext}}, W_{\text{int}}$) to $1.58$-bit representations (logically packed as $2$-bit structures in hardware). Notably, the Byte-level $p$-Spin model heavily reallocates its parameter budget from the vocabulary embedding projection into the core continuous logic matrices (Ternary MLP and State-Space Attention), reducing the global memory footprint to just $41.37$ MB while standard transformer or other natively uncompressed model would require ($116.62$ MB).

\begin{table*}[th]
\caption{Parameter breakdown for the standard Transformer Baseline.}
\label{tab:baseline_params}
\vskip 0.15in
\begin{center}
\begin{small}
\begin{sc}
\begin{tabular}{lr}
\toprule
Component & Parameter Count \\
\midrule
Total Parameters & $60,569,600$ \\
Embedding / Head (Tied) & $25,165,824$ \\
Non-Embedding Params & $35,403,776$ \\
\bottomrule
\end{tabular}
\end{sc}
\end{small}
\end{center}
\vskip -0.1in
\end{table*}

\begin{table*}[t]
\caption{Comprehensive Layer Capacity and Memory Footprint for the $p$-Spin Glass Sequence architectures (Subword and Byte variations). Natively quantized Ternary matrices strictly achieve an $8\times$ memory reduction compared to FP16 baselines.}
\label{tab:pspin_params}
\vskip 0.15in
\begin{center}
\begin{small}
\begin{sc}
\begin{tabular}{llrrr}
\toprule
Configuration & Layer Type & Params (M) & Logical Precision & Size (MB) \\
\midrule
\multirow{4}{*}{\textbf{Subword $p$-Spin}} 
 & Embeddings (Tied) & $26.74$ M & $32$-bit (FP32) & $102.00$ MB \\
 & Eq. MLP (Ternary) & $21.99$ M & $1.58$-bit (Packed 2b) & $5.24$ MB \\
 & State-Space Attention & $8.99$ M & $16$-bit (FP16/BF16) & $17.15$ MB \\
\cmidrule{2-5}
 & \textbf{Total / Footprint} & \textbf{57.72 M} & \textit{Uncompressed Eq: 161.09 MB} & \textbf{124.40 MB} \\
\midrule
\midrule
\multirow{4}{*}{\textbf{Byte $p$-Spin}} 
 & Embeddings (Tied) & $0.18$ M & $32$-bit (FP32) & $0.69$ MB \\
 & Eq. MLP (Ternary) & $45.09$ M & $1.58$-bit (Packed 2b) & $10.75$ MB \\
 & State-Space Attention & $15.69$ M & $16$-bit (FP16/BF16) & $29.93$ MB \\
\cmidrule{2-5}
 & \textbf{Total / Footprint} & \textbf{60.96 M} & \textit{Uncompressed Eq: 116.62 MB} & \textbf{41.37 MB} \\
\bottomrule
\end{tabular}
\end{sc}
\end{small}
\end{center}
\vskip -0.1in
\end{table*}

\bibliography{paper.bib}
\bibliographystyle{icml2025}
\end{document}